\documentclass[lettersize,journal]{IEEEtran}
\usepackage[T1]{fontenc}
\usepackage{amsmath,amsfonts}
\usepackage{algorithmic}
\usepackage{algorithm}
\usepackage{array}
\usepackage[caption=false,font=footnotesize]{subfig}
\usepackage{textcomp}
\usepackage{stfloats}
\usepackage{url}
\usepackage{verbatim}
\usepackage{graphicx}
\usepackage{cite}
\usepackage{booktabs}
\usepackage{makecell}
\usepackage[pagebackref, colorlinks=true, citecolor=blue, linkcolor=blue]{hyperref}
\def\ie{\emph{i.e.}} 

\def\eg{e.g.}

\def\etal{\emph{et al. }}

\begin{document}

\title{High Dynamic Range Video Reconstruction from \\ Single-Exposure Raw Sequences}

\author{Tao Zhang, Peixian Su, Xingyu Gao, Yunhao Zou, Yu Lu, Zunjie Zhu, Bolun Zheng, Ying Fu, Chenggang Yan
\thanks{Tao Zhang, Peixian Su, Yu Lu, Zunjie Zhu and Chenggang Yan are with School of Communication Engineering, Hangzhou Dianzi University, Hangzhou, China, and also with Zhejiang Provincial Key Laboratory of Low Altitude Ubiquitous Networking Technology, Hangzhou Dianzi University, Hangzhou, China.}
\thanks{Xingyu Gao is with the Institute of Microelectronics, Chinese Academy of Sciences, Beijing, China, and also with the University of Chinese Academy of Sciences, Beijing, China.}
\thanks{Yunhao Zou and Ying Fu are with School of Computer Science and Technology, Beijing Institute of Technology, Beijing, China.}
\thanks{Bolun Zheng is with School of Automation, Hangzhou Dianzi University, Hangzhou, China.}
\thanks{\textit{Corresponding author: Xingyu Gao.}}
}



\maketitle

\begin{abstract}

Due to the limited dynamic range of conventional image sensors, captured low dynamic range (LDR) video often suffers from highlight clipping and shadow detail loss, making high-quality high dynamic range (HDR) reconstruction from single-exposure sequences highly challenging without alternating exposures or extra hardware. Alternating-exposure HDR methods sacrifice frame rate and struggle with motion alignment, making them impractical for real-world capture. To address this, we propose RawHDRV, an end-to-end framework for single-exposure Raw video HDR reconstruction, that fundamentally exploits the linear response and channel-specific characteristics of Bayer data. Specifically, it features a channel-decomposition temporal alignment and fusion strategy that processes Bayer channels separately to exploit their distinct exposure characteristics, together with exposure-aware weighted fusion. It further incorporates an exposure complementarity mask-guided restoration module that leverages inter-frame exposure redundancy to adaptively fuse reliable information and suppress saturation artifacts, and introduces a mask-guided color loss that combines normalized error constraints with gradient smoothing to enhance highlight recovery. Furthermore, we construct a large-scale mobile Raw-HDR video dataset with per-frame HDR annotations. Experiments show that our method achieves the state-of-the-art results in all metrics, demonstrating superior spatial quality and temporal stability under extreme exposure conditions. The code is available at \url{https://github.com/supeixian/RawHDRV}. 
\end{abstract}

\section{Introduction}
\label{sec:Introduction}

\IEEEPARstart{C}{onstrained} by sensor sensitivity and bit depth, most consumer cameras often suffer from highlight clipping and shadow noise, which leads to the loss of important scene details~\cite{battiato2003high}. High dynamic range (HDR) imaging addresses this limitation by extending the recoverable luminance range while preserving both highlight and shadow information, and has become increasingly important in film production, mobile photography~\cite{hasinoff2016burst}, autonomous driving~\cite{shopovska2023high}, and virtual reality~\cite{matsuda2022realistic}.

Existing HDR reconstruction methods can be broadly divided into image-based and video-based paradigms. Early image HDR approaches~\cite{choi2020pyramid,endo2017deep,hasinoff2016burst,kalantari2017deep,yan2019attention,kong2024safnet} typically fuse multiple low dynamic range (LDR) images captured at different exposures to recover a wider luminance range. While effective for static scenes, these methods generally ignore temporal variation and therefore become unreliable in dynamic real-world scenarios. To avoid the capture and alignment difficulties of multi-exposure fusion, another line of research focuses on single-image LDR-to-HDR reconstruction, also known as inverse tone mapping~\cite{banterle2006inverse}, and has achieved substantial progress~\cite{yan2019multi,chen2021hdrunet,eilertsen2017hdr,santos2020single,xing2021invertible,zou2023rawhdr,chen2025hdr,kim2024dcdr,marnerides2018expandnet,wang2022kunet}.

\begin{figure}[t]
  \centering 
   \includegraphics[width=1\linewidth, trim=0cm 0.3cm 0cm 0cm, clip]{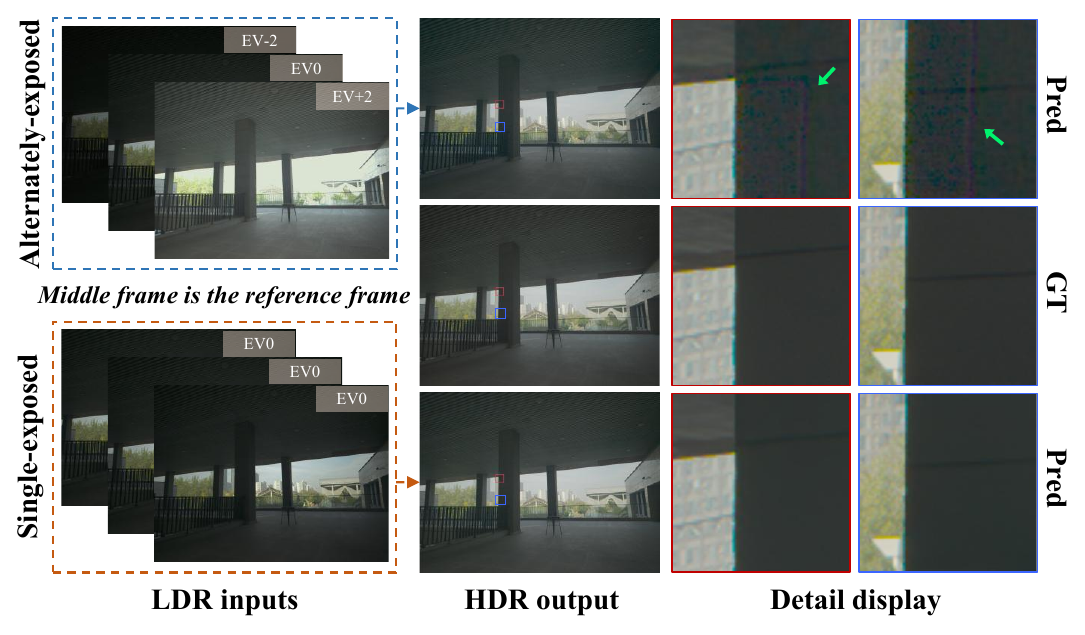}
   \caption{Alternating-exposure methods (Top) suffer from severe color fringing artifacts in challenging scenes from our RawHDRV dataset. In contrast, our single-exposure approach (Bottom) effectively avoids such artifacts and produces a more faithful HDR reconstruction.}
   \label{fig:onecol}
\end{figure}

To handle dynamic scenes, researchers have turned to video HDR reconstruction. Most existing video HDR methods adopt an alternating-exposure paradigm, in which frames captured at different exposure levels are fused across time~\cite{chen2021hdr,chung2023lan,kang2003high,mangiat2010high,shu2024towards,liu2023joint,cui2024exposure,xu2024hdrflow,guan2024diffusion,fang2019perceptual}. To reconstruct one HDR frame, these methods typically aggregate multiple LDR observations with varying exposures (\eg, a common $EV_{-2}, EV_{0}, EV_{+2}$ combination). However, this paradigm introduces two fundamental limitations in dynamic video scenes. First, mapping multiple captured frames to a single HDR output inevitably reduces the effective temporal resolution and enlarges the physical acquisition window. As a result, the output frame rate is lowered, and the inclusion of long-exposure frames makes the reconstruction more vulnerable to motion blur. Second, motion and exposure variation are strongly entangled across adjacent frames, making alignment inherently ill-posed: neighboring frames differ not only in geometry but also in brightness by several exposure values. This joint geometric-photometric mismatch often causes severe ghosting and color fringing in dynamic regions, as illustrated in Fig.~\ref{fig:onecol}. In contrast, a single-exposure HDR paradigm maintains constant exposure and a 1:1 temporal mapping with the sensor capture rate, thereby preserving temporal fidelity and reducing alignment to standard motion compensation. Table~\ref{tab:ae_vs_se_comparison} summarizes the key differences between the alternating-exposure and single-exposure paradigms.

\IEEEpubidadjcol

To alleviate the shortcomings of alternating exposures, several studies exploit event cameras~\cite{yang2023learning,wang2021asynchronous} or asynchronous dual-camera systems~\cite{zhang2025capturing} to reduce motion-induced artifacts or preserve exposure diversity. Although effective in specific settings, these solutions require specialized hardware and significantly increase system complexity and deployment cost.

Single-exposure video HDR reconstruction therefore offers a compelling alternative to alternating-exposure methods. However, it also poses two major challenges. First, without explicit exposure diversity, saturated highlights and noisy shadows cannot be directly complemented by differently exposed observations, and persistently over- or under-exposed regions lack sufficient luminance variation for reliable recovery. Second, most existing HDR methods~\cite{shu2024towards,chen2021hdr,chung2023lan,xu2024hdrflow} operate on ISP-processed sRGB inputs, which irreversibly discard critical radiometric information in extreme highlight and shadow regions. By contrast, Raw measurements preserve richer scene information owing to their linear sensor response and higher bit depth~\cite{zou2023rawhdr}. Nevertheless, the lack of large-scale Raw video datasets with per-frame HDR ground truth has greatly limited supervised learning for this problem, leaving single-exposure Raw video HDR reconstruction largely unexplored.

\begin{table*}[t]
  \centering
  \caption{Comparison between alternating-exposure and single-exposure video HDR paradigms. Single-exposure capture preserves full temporal resolution and avoids the motion-exposure misalignment inherent in alternating schemes.}
  \label{tab:ae_vs_se_comparison}
  \vspace{-0.4em}
  \small
  \renewcommand{\arraystretch}{1.15}
  \setlength{\tabcolsep}{3.5pt}

  \begin{tabular*}{\textwidth}{@{\extracolsep{\fill}}p{1.5cm} p{2.5cm} p{5.8cm} p{5.8cm}}
    \toprule
    \textbf{Category} & \textbf{Attribute} & \textbf{Alternating Exposure (Conventional)} & \textbf{Single-Exposure Raw (Ours)} \\
    \midrule
    \textbf{Temporal} & Input Sequence 
    & Varying exposures (\eg, $EV_{-2}, EV_{0}, EV_{+2}$). 
    & Constant exposure (\eg, $EV_{0}$). \\

    & Output Frame Rate 
    & \textbf{Reduced} ($F_{in}/3$), limited by the nominal $EV_{0}$ sample rate.  
    & \textbf{Full Rate} ($F_{in}$, 1:1 mapping). \\

    & Acquisition Window$^{\dagger}$ 
    & \textbf{Long}, including long exposure and thus more prone to motion blur. 
    & \textbf{Short}, better freezing high-speed motion. \\
    \midrule
    \textbf{Alignment} & Alignment Task 
    & \textbf{Ill-posed}. Requires aligning frames with drastic brightness differences ($2$--$4$ EV) and large displacements, often causing severe ghosting and color fringing. 
    & \textbf{Well-posed}. Standard motion compensation between frames with consistent brightness, thereby avoiding ghosting and color fringing. \\
    \bottomrule
  \end{tabular*}

  \begin{minipage}{\textwidth}
    \footnotesize
    \centering
    $^{\dagger}$\textit{Acquisition Window refers to the physical time span required to capture the data for a single HDR output frame.}
  \end{minipage}
\end{table*}

To address these challenges, we propose a single-exposure Raw-video HDR reconstruction framework that leverages temporal complementarity while explicitly modeling Bayer-channel characteristics. Specifically, we introduce a channel-aware parallel alignment and fusion strategy that processes the four Bayer channels (R, G1, G2, B) in separate branches to account for their distinct noise and saturation behaviors, enabling accurate detail recovery with preserved structural consistency. We further propose a mask-guided complementary fusion mechanism for highlight restoration, which adaptively integrates reliable inter-frame information while suppressing saturation artifacts, and design a mask-guided color loss to reinforce highlight recovery during training. In addition, we establish the first large-scale mobile Raw/HDR paired video dataset with per-frame HDR annotations, providing a reliable benchmark for single-exposure Raw-video HDR reconstruction. Extensive experiments on our dataset demonstrate that the proposed method achieves state-of-the-art performance.

Our contributions are summarized as follows:
\begin{itemize}
    \item We propose a channel-decomposition parallel alignment framework that models the four Bayer channels in parallel and leverages exposure-based masks for differential weighting, improving detail recovery and enhancing structural fidelity in extreme exposure regions.
    \item We introduce an exposure-complementarity mask-guided restoration scheme that partitions inter-frame regions and applies region-specific weighting to maximize reliable information transfer while suppressing saturation artifacts, enabling robust highlight recovery.
    \item We establish the first benchmark for single-exposure Raw-video HDR reconstruction by building a high-quality mobile Raw/HDR paired video dataset with per-frame annotations, enabling training and evaluation.
\end{itemize}

\section{Related Work}
\label{sec:Related Work}

We review the most relevant prior work, covering image-based HDR reconstruction, video-based HDR strategy, and Raw-domain processing with related datasets.

\subsection{HDR Image Reconstruction}

Multi-exposure fusion methods~\cite{choi2020pyramid,endo2017deep,hasinoff2016burst,kalantari2017deep,yan2019attention,kong2024safnet,hu2013hdr,sen2012robust,debevec2023recovering,niu2021hdr,tan2023deep} combine multiple LDR images captured at different exposures to expand the effective luminance range and achieve great success in static scenes. In contrast, reverse tone mapping reconstructs HDR from a single LDR image. 
Early approaches rely on illumination estimation or filter-based enhancement~\cite{banterle2006inverse,akyuz2007hdr,huo2014physiological}. With the development of deep learning~\cite{fu2021coded,fu2020joint,zhang2024survey}, more recent methods adopt end-to-end CNNs (\eg, ExpandNet~\cite{marnerides2018expandnet}, DeepHDR~\cite{santos2020single}, HDRUNet~\cite{chen2021hdrunet}, DCDR-UNet~\cite{kim2024dcdr}) to learn nonlinear mapping and improve highlight detail and global luminance restoration. 
Recent works further synthesize dense or pseudo multi-exposure stacks from a single LDR image and fuse them to leverage the benefits of multi-exposure fusion (\eg, Lee~\etal\cite{lee2018deep}, Chen~\etal\cite{chen2023learning}).

Some HDR approaches employ specialized hardware, such as event cameras~\cite{han2020neuromorphic,messikommer2022multi} or infrared guidance~\cite{zhao2024equivariant,peng2025hdrt}, to capture richer dynamic range or motion cues. Recently, Zou \etal\cite{zou2023rawhdr} show that high-bit-depth Raw data preserving linear, unprocessed sensor responses enables effective single-image HDR reconstruction.

\subsection{HDR Video Reconstruction}

Hardware-based HDR video capture typically relies on elaborate optics, such as scanning-exposure/sensitivity systems~\cite{heide2014flexisp,choi2017reconstructing} or internal/external beam splitters~\cite{kronander2014unified,mcguire2007optical,tocci2011versatile}, which yield high-quality HDR video but are costly and complex to build. Consequently, reconstructing HDR video from LDR inputs emerges as a practical alternative.

Existing HDR video methods~\cite{chen2021hdr,chung2023lan,shu2024towards,xu2024hdrflow} primarily rely on alternating-exposure sequences to expand dynamic range. Early approaches use optical flow~\cite{kang2003high} or block-based motion estimation~\cite{mangiat2010high} for alignment before fusion. With deep learning, learned flow networks, deformable convolutions, and adaptive weighting modules are adopted to improve alignment and fusion accuracy~\cite{chen2021hdr,kalantari2019deep,lee2018deep,zhang2025unaligned}. However, large inter-frame brightness differences still cause challenging registration and ghosting under complex motion~\cite{chung2023lan}. To address this, Shu \etal\cite{shu2024towards} propose a two-stage alignment strategy for dynamic scenes, while Xu \etal\cite{xu2024hdrflow} design a multi-scale flow network with large kernels to better handle large motions.

To circumvent the inherent limitations of alternating-exposure schemes, we propose a new paradigm, \ie, HDR video reconstruction under single-exposure conditions. This task leverages temporal redundancy from consistently exposed consecutive frames to enhance per-frame dynamic range and structural consistency, a setting closely related to multi-frame fusion in video super-resolution~\cite{wang2019edvr,zhang2024realviformer,yue2022real} and video restoration~\cite{li2023simple,liang2024vrt}, where optical flow or deformable convolution enables precise alignment and temporal aggregation. These advances highlight the potential of temporal complementarity to overcome single-frame information deficits, providing a strong foundation for single-exposure HDR video reconstruction.

\subsection{Raw Processing and Video HDR Dataset}

Most HDR reconstruction methods rely on ISP-processed sRGB data, but the ISP’s nonlinear mapping, white balancing, and quantization inevitably discard highlight and shadow information~\cite{zou2023rawhdr}. Early work by Kalantari \etal\cite{kalantari2013patch} captures 5 LDR sequences using two alternating exposure strategies. Chen \etal\cite{chen2021hdr} use alternating exposures to capture a dataset of 76 dynamic image pairs, 49 static pairs, and 50 unlabeled sequences for HDR video research. Later, Shu \etal\cite{shu2024towards} collect a large-scale dataset where each frame contains multi-exposure inputs and HDR labels, advancing sRGB-based video HDR research.

In contrast, Raw sensor data, thanks to its linear response and higher bit depth (10-16 bits), preserves richer radiometric information \cite{zhang2022deep,zhang2024deep,fu2022low,fu2023raw,zhang2022guided} and offers a promising path to overcome the limits of single-exposure HDR. Recent efforts begin to explore Raw-domain HDR. Zou \etal\cite{zou2023rawhdr} release high-quality Raw-HDR pairs in single-images, demonstrating the potential of Raw data. Yue \etal\cite{yue2023hdr} collect 85 mobile Raw LDR-HDR video pairs, though still using alternating exposures. Despite their value, publicly available Raw video datasets with per-frame HDR ground truth under single-exposure capture remain scarce, severely limiting supervised learning and fair benchmarking for single-exposure Raw-video HDR reconstruction.
\section{RawHDRV Network}
\label{sec:Method}

\begin{figure}[t]
\centering
\subfloat[CRF]{
\includegraphics[width=0.46\linewidth]{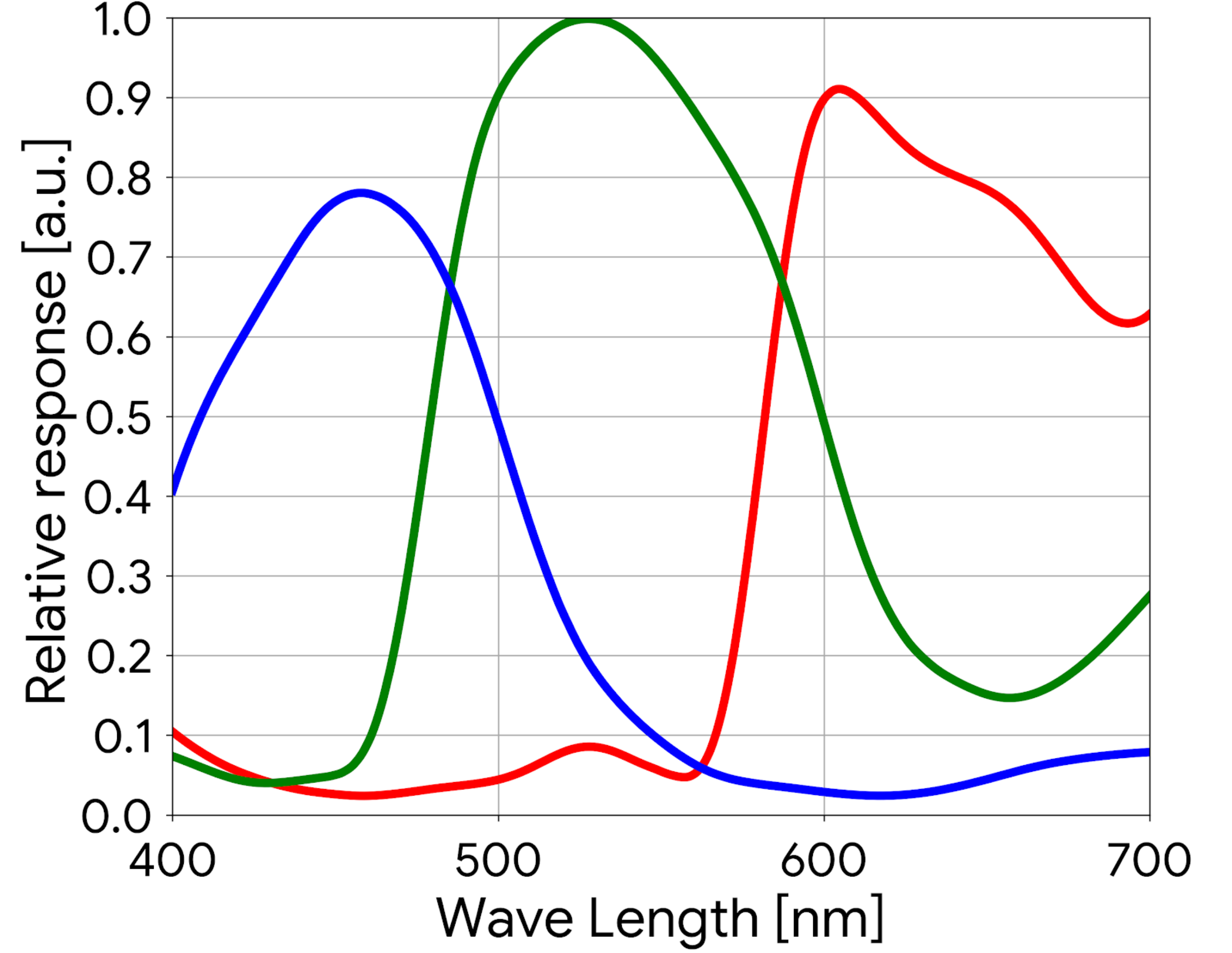}
\label{fig:motivation-a}}
\hfil
\subfloat[Per-channel Mean]{
\begin{minipage}[t]{0.46\linewidth}
\centering
\vspace{-28mm}
\renewcommand{\arraystretch}{1.1}
\setlength{\tabcolsep}{6pt}
\small
\begin{tabular}{c|c}
\toprule
\textbf{Channel} & \textbf{Mean Value} \\
\midrule
Red   & 31.36 \\
\midrule
Green & 64.39 \\
\midrule
Blue  & 33.49 \\
\bottomrule
\end{tabular}
\end{minipage}
\label{fig:motivation-b}}

\subfloat[Examples of exposure complementarity across frames.]{
\includegraphics[width=\linewidth]{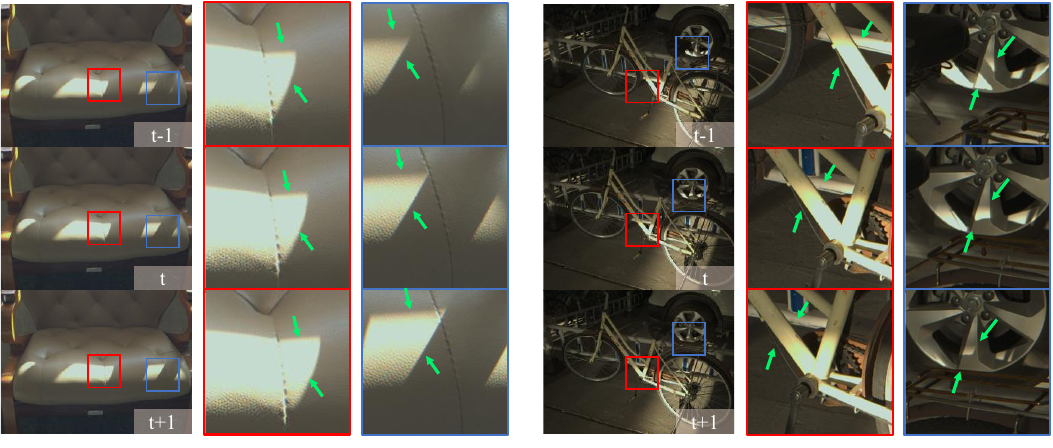}
\label{fig:motivation-c}}

\caption{
Motivation visualization.
(a) Camera response function of a typical Sony IMX-series sensor.
(b) Channel-wise mean intensity statistics under standard exposure (EV0).
(c) Exposure complementarity across adjacent frames.
}
\label{fig:motivation}
\end{figure}

\begin{figure*}[t]  
  \centering
  \includegraphics[width=1\textwidth, trim=0cm 0cm 0cm 0.23cm, clip]{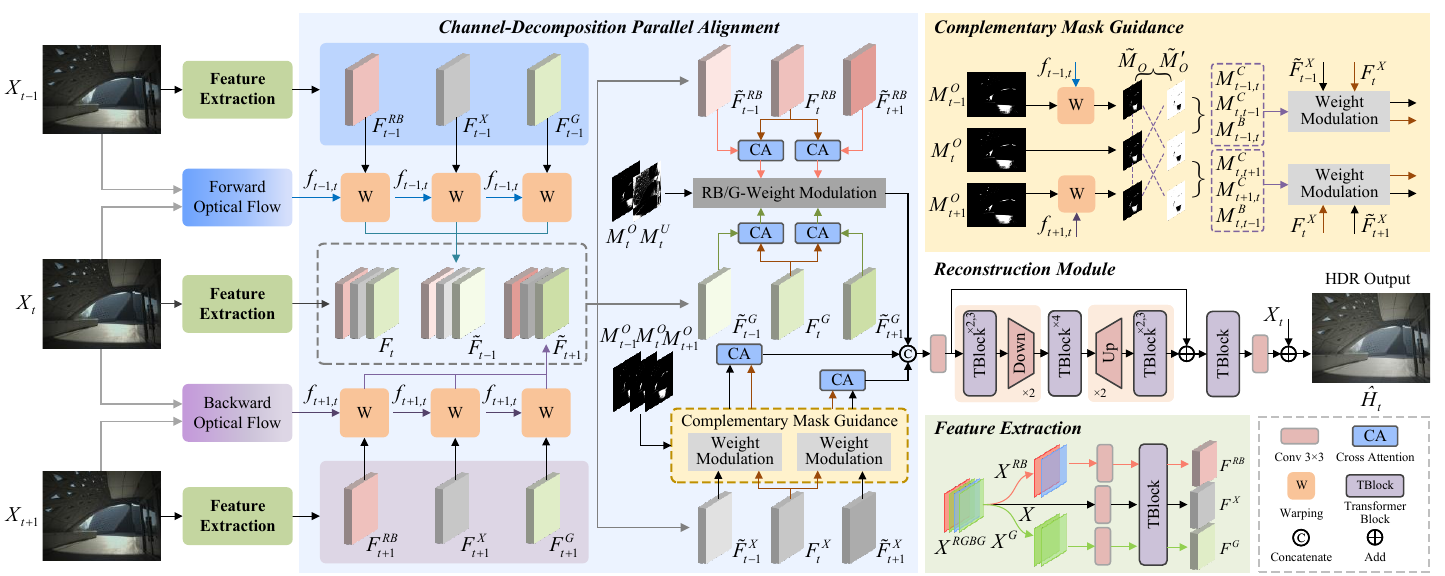}
  \caption{Overview of the proposed RawHDRV architecture, integrating channel-decomposition parallel alignment and exposure complementarity mask-guided restoration applied to the full input sequence.}
  \label{fig:model_architecture}
\end{figure*}

This section outlines our problem formulation and motivation for single-exposure Raw-to-HDR video reconstruction, and the network architecture of our proposed RawHDRV model. The overall framework is illustrated in Fig.~\ref{fig:model_architecture}.

\subsection{Problem Formulation and Motivation}
Given a single-exposure Raw video sequence $\{ I_t \in \mathbb{R}^{4 \times H \times W} \}_{t=0}^{n-1}$ encoded in the RGBG (four-channel Bayer mosaic) format, our goal is to reconstruct the corresponding HDR video sequence $\{ H_t \in \mathbb{R}^{4 \times H \times W} \}_{t=0}^{n-1}$. Let $f(\cdot\,; \theta)$ represent a neural network parameterized by $\theta$. The reconstruction process is formulated as
\begin{equation}
H_t = f(I_t; \theta).
\end{equation}

Reconstructing HDR content from such single-exposure Raw sequences is highly challenging. The limited sensor dynamic range causes irreversible highlight clipping and shadow noise, and unlike alternating-exposure methods~\cite{chen2021hdr,chung2023lan,cui2024exposure}, the absence of exposure diversity forces reliance on temporal redundancy, yet persistent over- or under-exposure often leaves critical regions hard to recover.

Typical Bayer sensors exhibit significant channel-wise differences \cite{zou2023rawhdr,zhang2024deep}, \ie, green pixels possess higher photon capture efficiency and signal-to-noise ratio. It provides reliable detail shadows but saturate easily in highlights, whereas red and blue channels retain more highlight information at the cost of increased noise in dark regions (Fig.~\ref{fig:motivation}(a)(b)). It implies that subchannels offer complementary cues under varying exposure. Moreover, even under fixed exposure, temporal exposure complementarity arises in video. Due to scene or camera motion (\eg, translation, rotation, or sub-pixel shifts), a pixel saturated in one frame may be well-exposed in a neighboring frame (Fig.~\ref{fig:motivation}(c)).

Motivated by these two priors, channel-specific characteristics and inter-frame exposure complementarity, we propose a channel-decomposition temporal alignment and fusion mechanism together with an exposure-complementary, mask-guided restoration module, which jointly enhance highlight recovery and HDR fidelity by leveraging reliable information across channels and frames.

\subsection{Model Architecture}
\noindent\textbf{Network Overview.}
Our network takes three consecutive Raw frames $\{X_{t-1}, X_t, X_{t+1}\}$ to predict the HDR output $\hat{H}_t$ (Fig.~\ref{fig:model_architecture}). It first extracts and aligns channel-specific features from the Bayer mosaic through a \textit{channel-decomposition and parallel alignment} strategy, followed by exposure-aware fusion. An \textit{Exposure Complementarity Mask-Guided Restoration} module then leverages temporal exposure complementarity to recover overexposed details. Finally, the multi-channel, bidirectionally enhanced features are fused and refined through a multi-scale encoder-decoder, with a residual connection yielding an HDR frame with extended dynamic range and richer detail.

\vspace{0.5em}
\noindent\textbf{Channel-Decomposition Parallel Alignment.}
Inspired by RawHDR~\cite{zou2023rawhdr}, we propose a channel-decomposition framework that exploits the distinct noise and saturation behaviors of Bayer Raw channels by splitting the RGGB input into three parallel streams: (1) the $X$ stream (all four channels) for global context and structural consistency, (2) the $RB$ stream (Red and Blue) for highlight detail recovery, and (3) the $G$ stream (dual Green channels) for dark-area denoising and detail preservation.

For each frame $X_i$ ($i \in \{t-1, t, t+1\}$), we extract shallow features per stream
\begin{equation}
F_i^k = E_k(X_i^k), \quad k \in \{X, RB, G_1G_2\},
\end{equation}
where $E_k$ consists of convolution and Transformer~\cite{zhang2024realviformer} blocks. Optical flow $\mathbf{f}_{t \pm 1 \to t}$ is estimated on the corresponding sRGB frames using SPyNet~\cite{ranjan2017optical}. The resulting flow is then applied to warp the Raw-domain features of adjacent frames toward the center frame
\begin{equation}
\tilde{F}_{t \pm 1}^k = \mathcal{W}(F_{t \pm 1}^k, \mathbf{f}_{t \pm 1 \to t}), \quad k \in \{X, RB, G_1G_2\}.
\end{equation}

To mitigate alignment errors from optical-flow residuals, we apply a cross-frame channel cross-attention module ($CA$) independently to the $RB$ and $G$ streams
\begin{equation}
A_{t \pm 1}^s = CA(F_t^s, \tilde{F}_{t \pm 1}^s), \quad s \in \{RB, G_1G_2\},
\end{equation}
which preserves channel-specific characteristics and local exposure differences.

We then compute soft overexposure ($M_t^O$) and underexposure ($M_t^U$) masks for the center frame by thresholding Raw intensities ($>$ 0.95 for overexposure, $<$ 0.05 for underexposure) followed by Gaussian blur. Let $M_t^N = 1 - M_t^O - M_t^U$ denote the normal region mask. Pixel-wise adaptive weights are applied to the $RB$ and $G$ streams
\begin{equation}
\begin{split}
w_{RB} &= \alpha_1 \cdot M_t^O + \alpha_2 \cdot M_t^N + \alpha_3 \cdot M_t^U, \\
w_{G}  &= \alpha_1 \cdot M_t^U + \alpha_2 \cdot M_t^N + \alpha_3 \cdot M_t^O,
\end{split}
\label{eq:weights}
\end{equation}
with $\alpha_1 = 2.0$, $\alpha_2 = 1.0$, $\alpha_3 = 0.3$. Weighted features are
\begin{equation}
\hat{F}_{t \pm 1}^{RB} = w_{RB} \odot A_{t \pm 1}^{RB}, \quad 
\hat{F}_{t \pm 1}^{G} = w_{G} \odot A_{t \pm 1}^{G},
\end{equation}
where $\odot$ denotes element-wise multiplication. It realizes differentiated channel weighting to enhance $RB$ in highlights, boost $G$ in shadows, and balance contributions in normal regions.

\vspace{0.5em} 
\noindent\textbf{Exposure Complementarity Mask-Guided Restoration.}
Based on the observation that overexposed regions often differ spatially across adjacent frames due to motion, we design a mask-guided restoration module to recover saturated details in the center frame by leveraging reliable information from neighbors.

First, we compute overexposure masks $M_i^O$ for each frame $i \in \{t-1, t, t+1\}$, and warp the neighboring masks to the center frame using optical flow
\begin{equation}
\tilde{M}_{t-1}^O = \mathcal{W}(M_{t-1}^O, \mathbf{f}_{t-1 \to t}).
\end{equation}
By comparing the center mask $M_t^O$ with the warped mask $\tilde{M}_{t-1}^O$, we binarize them ($M^{\text{bin}} = \mathbb{I}(M^O > 0.5)$) and partition the spatial domain into three mutually exclusive regions
\begin{equation}
\begin{split}
M_{t-1,t}^C &= (1 - M_t^{\text{bin}}) \odot \tilde{M}_{t-1}^{\text{bin}}, \\
M_{t,t-1}^C &= M_t^{\text{bin}} \odot (1 - \tilde{M}_{t-1}^{\text{bin}}), \\
M_{t-1,t}^B &= M_t^{\text{bin}} \odot \tilde{M}_{t-1}^{\text{bin}},
\end{split}
\label{eq:region_masks}
\end{equation}
where $M_{t-1,t}^C$ and $M_{t,t-1}^C$ denote complementary regions (one overexposed, one normal), the most valuable for restoration, as the neighboring frame retains valid brightness and texture at the same location.

We then apply a rule-based weighting strategy to the center and propagated features $(F_t^X, \tilde{F}_{t-1}^X)$
\begin{equation}
\begin{split}
w_t &= 1 + (\alpha - 1) M_{t-1,t}^C - (1 - \beta) \, \Gamma_t, \\
w_{t-1} &= 1 + (\alpha - 1) M_{t,t-1}^C - (1 - \beta) \, \Gamma_{t-1},
\end{split}
\label{eq:fusion_weights}
\end{equation}
with $\Gamma_t = M_{t,t-1}^C + M_{t-1,t}^B$, $\Gamma_{t-1} = M_{t-1,t}^C + M_{t-1,t}^B$, and $\alpha = 5.0$, $\beta = 0.1$. The enhanced features are obtained as
\begin{equation}
    \hat{F}_t^X = w_t \odot F_t^X, \quad \hat{F}_{t-1}^X = w_{t-1} \odot \tilde{F}_{t-1}^X.
\end{equation}
This design enhances normally exposed frames in complementary regions, suppresses overexposed interference, and attenuates both frames in doubly overexposed areas to prevent error propagation. The enhanced features are fused via cross-frame attention to produce $A_{t-1}^X$.

The process is performed bidirectionally (forward: $t-1 \to t$, backward: $t+1 \to t$), yielding $A_{t-1}^X$ and $A_{t+1}^X$. These are concatenated with the weighted RB and G features and integrated via a $3\times3$ convolution
\begin{equation}
F_{\text{fuse}} = \text{Conv}\big([A_{t-1}^X, A_{t+1}^X, \hat{F}_{t \pm 1}^{RB}, \hat{F}_{t \pm 1}^{G}]\big),
\end{equation}
where $[\cdot]$ denotes channel-wise concatenation.

\vspace{0.5em} 
\noindent\textbf{Fusion and Reconstruction.}
The fused features $F_{\text{fuse}}$ are processed by a multi-scale encoder-decoder. The encoder uses three downsampling levels with $[2, 3, 4]$ stacked Transformer Blocks to capture multi-scale context, while the symmetric decoder employs skip connections to preserve fine structural details. The output is refined through a refinement module and a convolutional layer, and the final HDR raw frame is reconstructed via a residual connection with the center input
\begin{equation}
\hat{H}_t = \text{Conv}(T(F_{\text{fuse}})) + X_t,
\end{equation}
where $T(\cdot)$ denotes the encoder-decoder and refinement pipeline. This design ensures simultaneous recovery of extended dynamic range and high-fidelity texture details.

\subsection{Loss Function}
To achieve accurate HDR video reconstruction, we use three complementary losses. \(L_1\) loss ensures overall brightness consistency, while log-\(L_2\) loss mitigates highlight dominance in the linear domain by operating in the logarithmic space
\begin{equation}
\mathcal{L}_{\text{log-L2}} = \|\log(\hat{H}_t + \epsilon) - \log(H_t + \epsilon)\|_2^2,
\end{equation}
with \(\epsilon = 1/5000\), balancing dark and bright regions to improve mid-tone detail recovery.

\vspace{0.5em} 
\noindent\textbf{Mask-Guided Color Loss.}
To enhance highlight recovery, we introduce a Mask-Guided Color Loss that concentrates supervision on overexposed regions using mask \(M_t^O\). The loss comprises two terms:  
(i) a \textit{mask-normalized} \(L_1\) loss that prioritizes severely clipped pixels by averaging the masked error over the overexposed area, and  
(ii) a \textit{gradient smoothness} term that enforces spatial continuity and reduces artifacts at highlight boundaries. Formally,
\begin{equation}
\begin{split}
\mathcal{L}_{\text{m}} = \lambda_o \cdot \mathbb{E}_{b,c} \left[ \frac{ \langle |\hat{H}_{b,c} - H_{b,c}|, M_b \rangle }{ \|M_b\|_1 + \tau } \right] \\
+ \lambda_s \left( \|\nabla_x \hat{H}_t\|_1 + \|\nabla_y \hat{H}_t\|_1 \right),
\end{split}
\end{equation}
where \(\langle \cdot, \cdot \rangle\) denotes pixel-wise inner product, \(\|M_b\|_1\) is the total mask area, and \(\tau = 10^{-8}\) ensures numerical stability. With \(\lambda_o = 2.0\) and \(\lambda_s = 0.1\), this loss improves PSNR-$\mu$ in overexposed regions by \(\sim\)1.1 dB and yields visually smoother, more natural results.

The final total loss function is defined as
\begin{equation}
\mathcal{L}=\mathcal{L}_{L1} + \mathcal{L}_{\text{log-L2}} + \lambda \mathcal{L}_{\text{m}},
\end{equation}
where the Mask-Guided Color Loss weight $\lambda = 0.3$.

\section{The Proposed RawHDRV Dataset}
\label{sec:Proposed Dataset}

\begin{figure*}[h]
  \centering
  \includegraphics[width=1\textwidth]{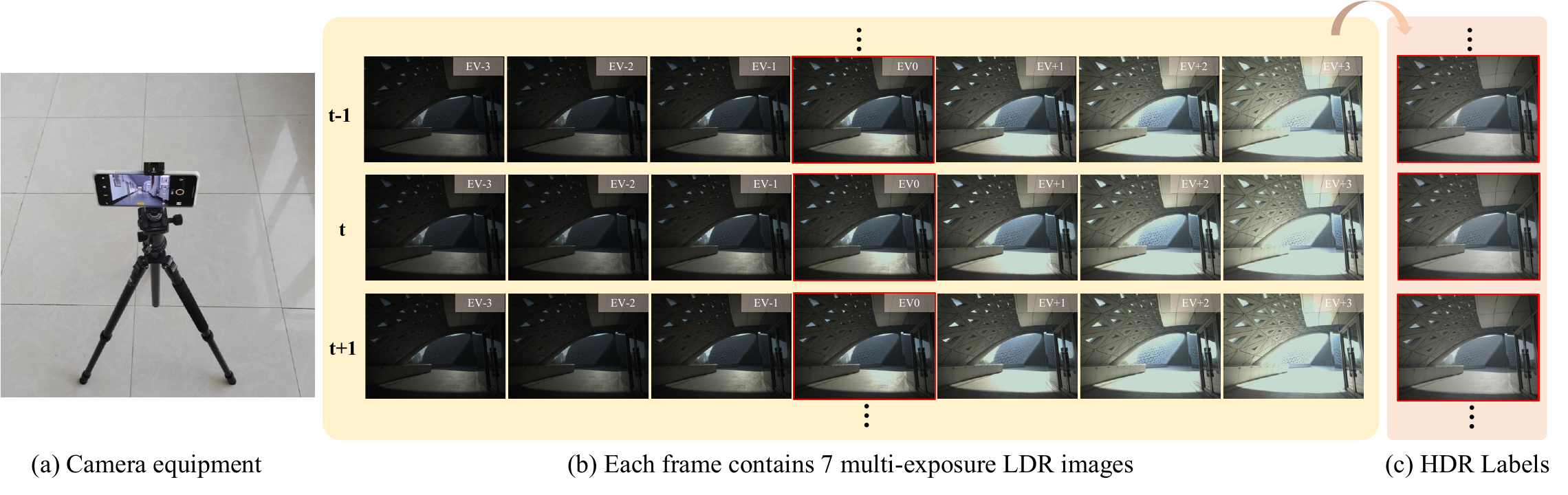}
  \vspace{-2.2em}
  \caption{Data acquisition and ground truth pipeline. 
(a) Tripod-based mobile capture setup. 
(b) Frame-level bracketing protocol: a 7-exposure stack ($\text{EV} \in [-3, +3]$) is captured per timestamp, where the standard exposure ($\text{EV}0$, red box) serves as the input. 
(c) Per-frame HDR Ground Truth, generated via multi-exposure fusion and radiometrically calibrated to the $\text{EV}0$ photometric baseline.}
\vspace{-0.5em}
  \label{fig:data_acquisition}
\end{figure*}

\begin{figure*}[t]
\centering

\subfloat[Motion types]{
\includegraphics[width=0.25\textwidth]{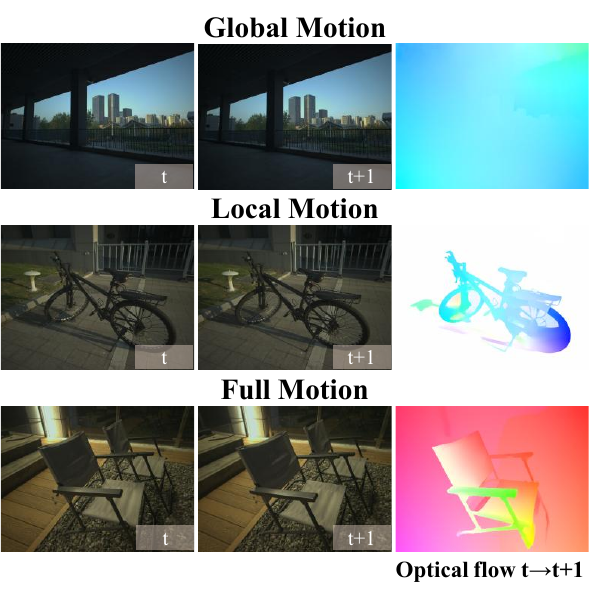}
\label{fig:short-a}}
\hfil
\subfloat[Statistics]{
\includegraphics[width=0.25\textwidth]{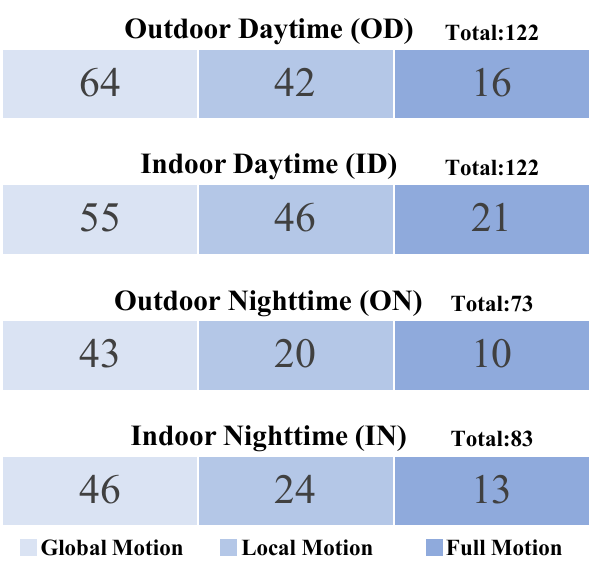}
\label{fig:short-b}}
\hfil
\subfloat[Dataset scenes]{
\includegraphics[width=0.45\textwidth]{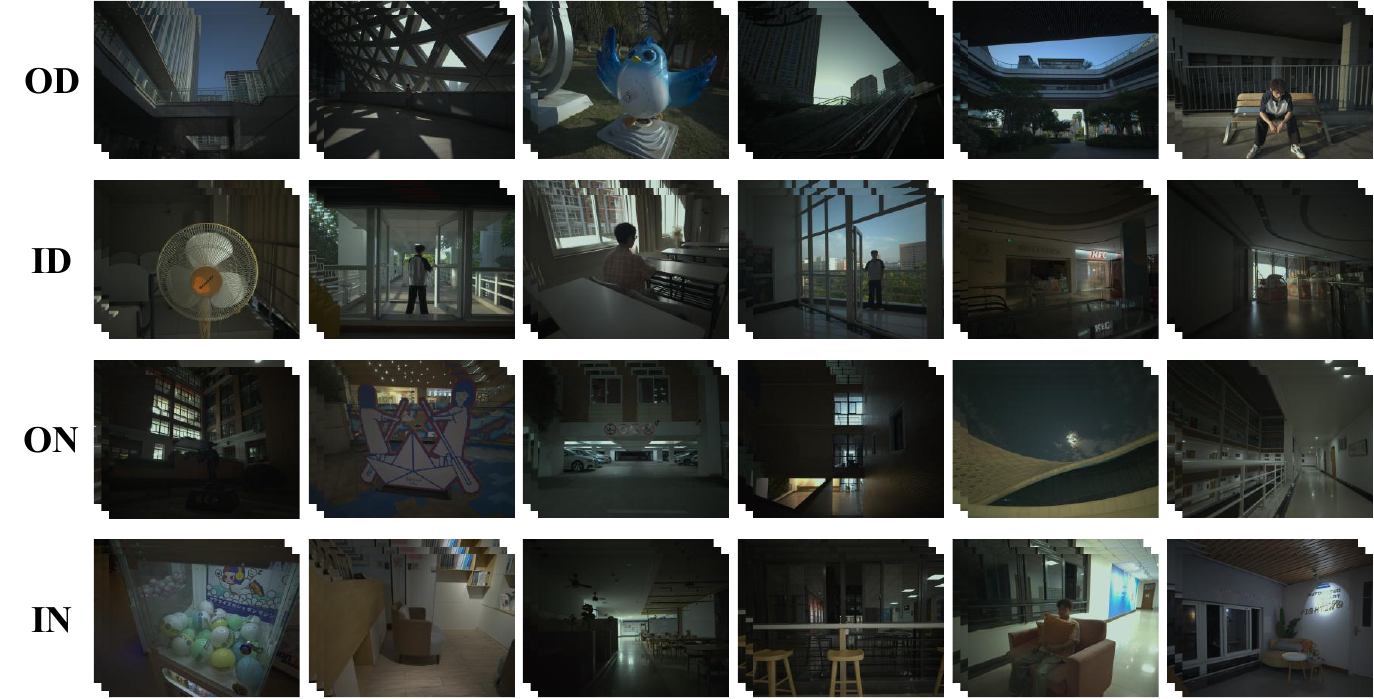}
\label{fig:short-c}}

\caption{(a) Illustration of three motion types: global motion, local motion, and full motion. The optical flow between adjacent frames is computed using RAFT~\cite{teed2020raft}. (b) Scene and motion distribution of the dataset. (c) Typical scenes categorized into four types: outdoor daytime (OD), indoor daytime (ID), outdoor nighttime (ON), and indoor nighttime (IN).}
\label{fig:short}
\end{figure*}

\begin{table*}[t]
  \centering
  \vspace{-0.2em}
  \caption{Comparison of representative HDR and Raw-domain datasets.}
  \vspace{-0.5em}
  \label{tab:hdr_datasets}
  \resizebox{\textwidth}{!}{
  \begin{tabular}{lcccccc}
    \toprule
    \textbf{Dataset} & \textbf{Year} & \textbf{Scenes} & \textbf{Exposure Type} & \textbf{Resolution} & \textbf{Format / Bit-depth} & \textbf{Applicable Tasks} \\
    \midrule
    Kalantari \etal~\cite{kalantari2013patch} & 2013 & 5 & Alternating & 1280$\times$720 & sRGB / 8-bit & Multi-exposure HDR video \\
    Chen \etal~\cite{chen2021hdr} & 2021 & 175 & Alternating & 1536$\times$813 & sRGB / 8-bit & Multi-exposure HDR video \\
    Zou \etal~\cite{zou2023rawhdr} & 2023 & 324 & Single & 6720$\times$4480 & Raw / 14-bit & Single-image Raw-to-HDR \\
    Yue \etal~\cite{yue2023hdr} & 2023 & 85 & Alternating & 4000$\times$3000 & Raw / 10-bit & Multi-exposure HDR video \\
    Shu \etal~\cite{shu2024towards} & 2024 & 500 & Multiple & 1500$\times$1000 & sRGB / 8-bit & sRGB-to-HDR video \\
    \textbf{RawHDRV (Ours)} & \textbf{2025} & \textbf{400} & \textbf{Multiple} & \textbf{4096$\times$3072} & \textbf{Raw / 10-bit} & \textbf{Raw-to-HDR video} \\
    \bottomrule
  \end{tabular}
  }
  \par\vspace{0.3ex}
  \begin{minipage}{\textwidth}
    \footnotesize
    \centering
    \textit{Note: Dataset format refers to the input format used during training in the corresponding paper.}
  \end{minipage}
\end{table*}

\begin{table}[tbp]
\centering
\caption{Motion magnitudes (mean optical flow in pixels).}
\vspace{-0.5em}
\label{tab:motion_dist}
\renewcommand{\arraystretch}{0.9}
\resizebox{0.98\columnwidth}{!}{
\begin{tabular}{@{}l|cccc|c@{}}
\toprule
Magnitude (px) & [0, 2) & [2, 10) & [10, 30) & $\ge$ 30 & Total \\ \midrule
Count & 1,780,255 & 334,587 & 3,093,350 & 1,104,768 & 6,312,960 \\
Percentage (\%) & 28.15\% & 5.29\% & 48.99\% & 17.57\% & 100.0\% \\
\bottomrule
\end{tabular}
}
\end{table}

To support research on single-exposure Raw-video HDR reconstruction, we present RawHDRV, a large-scale mobile Raw video dataset with per-frame HDR ground truth. Following the capture protocol of Real-HDRV~\cite{shu2024towards} and extending it to mobile raw video, we use a device equipped with a Sony IMX989 sensor to capture, for each video frame, seven LDR images in rapid succession under exposure values $\text{EV} \in \{-3, -2, -1, 0, +1, +2, +3\}$. Each LDR image is saved as a 10-bit Bayer Raw file at $3072 \times 4096$ resolution. A radiometric HDR ground truth is generated per frame via a classic multi-exposure fusion algorithm~\cite{debevec2023recovering}. The mobile device and a typical 7-exposure sequence are visualized in Fig.~\ref{fig:data_acquisition}.

We collect data across four illumination conditions (indoor/outdoor, day/night) and three motion patterns to ensure diversity: (1) \textit{global motion}, where the camera moves while the scene is static; (2) \textit{local motion}, with the camera fixed and foreground objects moving; and (3) \textit{full motion}, where both camera and foreground move to simulate complex real-world dynamics (Fig.~\ref{fig:short}(a)). All sequences are captured on a tripod and triggered via wireless remote to eliminate hand tremor. We discard samples with failed exposure bracketing or misalignment, and log key metadata, including white balance and color correction matrices, which are stored in the HDR files.

The final dataset contains 400 video sequences (244 daytime, 156 nighttime), each with 5 to 8 consecutive frames, yielding 2,455 aligned Raw-HDR pairs. Fig.~\ref{fig:short}(b)(c) summarize the scene and motion distribution and show representative samples. We performed patch-level optical flow analysis using RAFT~\cite{teed2020raft} on ISP-processed images at half the original Raw resolution (2048$\times$1536 vs.\ 4096$\times$3072). Results in Table~\ref{tab:motion_dist}, 28.15\% of patches exhibit near-static motion ($<2$ pixels displacement), while 66.6\% of patches contain significant motion ($\geq10$ pixels displacement), including 17.6\% with large motion ($\geq30$ pixels displacement). This distribution confirms our dataset contains both substantial low-motion regions (enabling reliable temporal information borrowing) and highly dynamic scenarios (posing challenging alignment conditions).

To contextualize our contribution, we summarize representative HDR and raw-domain datasets in Table~\ref{tab:hdr_datasets}. Existing works primarily focus on either sRGB-based multi-exposure HDR video~\cite{kalantari2013patch,chen2021hdr,shu2024towards} or single-image raw-to-HDR tasks~\cite{zou2023rawhdr}. Notably, Yue~\etal~\cite{yue2023hdr} introduced a raw video dataset, but it still relies on alternating exposures. In contrast, our RawHDRV is the first large-scale raw video dataset specifically designed for \textit{single-exposure} HDR reconstruction, offering per-frame HDR ground truth and diverse real-world scenes.

Beyond single-exposure Raw-video HDR, RawHDRV also supports multi-exposure video HDR~\cite{shu2024towards}, multi-exposure image fusion~\cite{liu2023joint}, and single-image Raw HDR~\cite{zou2023rawhdr}. We will publicly release the dataset and code upon paper acceptance.

\begin{table*}[t]
  \centering
  \caption{Comparison of HDR reconstruction performance on our dataset. The upper block lists image-based methods (applied frame-wise), and the lower block lists video-based methods (using 3-frame input). Best results are highlighted in \textbf{bold}.}
  \vspace{-0.5em} 
  \label{tab:comparison}
  \small
  \setlength{\tabcolsep}{4pt} 
  \resizebox{1\textwidth}{!}{%
  \begin{tabular}{l|cccccc|ccc}
    \toprule
    \textbf{Model}& \textbf{PSNR-L$\uparrow$} & \textbf{PSNR-$\mu$$\uparrow$} & \textbf{SSIM-L$\uparrow$} & \textbf{MS-SSIM$\uparrow$} & \textbf{HDR-VDP-2$\uparrow$} & \textbf{HDR-VQM$\downarrow$} & \textbf{Params(M)$\downarrow$} & \textbf{Flops(T)}$\downarrow$ & \textbf{Time(s)}$\downarrow$ \\
    \midrule
    ExpandNet~\cite{marnerides2018expandnet} & 34.78 & 26.29 & 0.9795 & 0.9813 & 69.20 & 0.0217 & 0.46 & 0.108 & 0.01 \\ 
    DeepHDR~\cite{santos2020single} & 38.32 & 30.26 & 0.9859 & 0.9805 & 69.69 & 0.0145 & 51.55 & 0.153 & 0.01 \\ 
    HDRUNet~\cite{chen2021hdrunet} & 41.37 & 34.52 & 0.9909 & 0.9870 &  71.09 & 0.0110 & 1.65 & 0.187 & 0.02 \\ 
    KUNet~\cite{wang2022kunet} & 38.56 & 33.13 & 0.9823 & 0.9864 & 70.47 & 0.0177 & 1.14  & 0.338 & 0.02 \\ 
    RawHDR~\cite{zou2023rawhdr} & 38.38 & 32.00 & 0.9835 & 0.9848 & 70.06 & 0.0169 & 10.48 & 0.326 & 0.05 \\ 
    RECNet~\cite{liu2024region} & 36.59 & 27.10 & 0.9846 & 0.9812 & 68.48 & 0.0187 & 1.71 & 0.824 & 0.20 \\ 
    MaIR~\cite{li2025mair} & 39.15 & 32.27 & 0.9885 & 0.9853 & 70.61 & 0.0130 & 1.48 & 0.651 & 0.89 \\ 
    \midrule
    EDVR~\cite{wang2019edvr} & 37.19 & 26.73 & 0.9850 & 0.9735 & 69.22 & 0.0185 & 3.03 & 1.997 & 0.18 \\ 
    Shift-Net~\cite{li2023simple} & 37.70 & 29.27 & 0.9874 & 0.9847 & 70.45 & 0.0147 & 12.99 & 3.068 & 0.58 \\ 
    DSTNet~\cite{pan2023deep} & 39.74 & 32.57 & 0.9874 & 0.9797 & 69.83 & 0.0128 & 7.45 & 1.137 & 0.07 \\ 
    RealViformer~\cite{zhang2024realviformer} & 41.60 & 35.38 & 0.9910 & 0.9876 & 70.99 & 0.0102 & 8.69 & 2.870 & 0.36 \\ 
    VRT~\cite{liang2024vrt} & 40.33 & 32.82 & 0.9894 & 0.9833 & 70.05 & 0.0112 & 2.54 & 0.994 & 0.68 \\ 
    RViDeformer~\cite{yue2025rvideformer} & 37.27 & 28.23 & 0.9868 & 0.9811 & 69.82 & 0.0156 & 0.66 & 0.412 & 1.08 \\ 
    \midrule
    \textbf{Ours} & \textbf{44.14} & \textbf{37.77} & \textbf{0.9941} & \textbf{0.9916} & \textbf{72.45} & \textbf{0.0066}  & 8.86   & 1.912  & 0.25 \\
    \bottomrule
  \end{tabular}
  }
\end{table*}

\begin{figure*}[t]
\centering
  \includegraphics[width=1\textwidth]{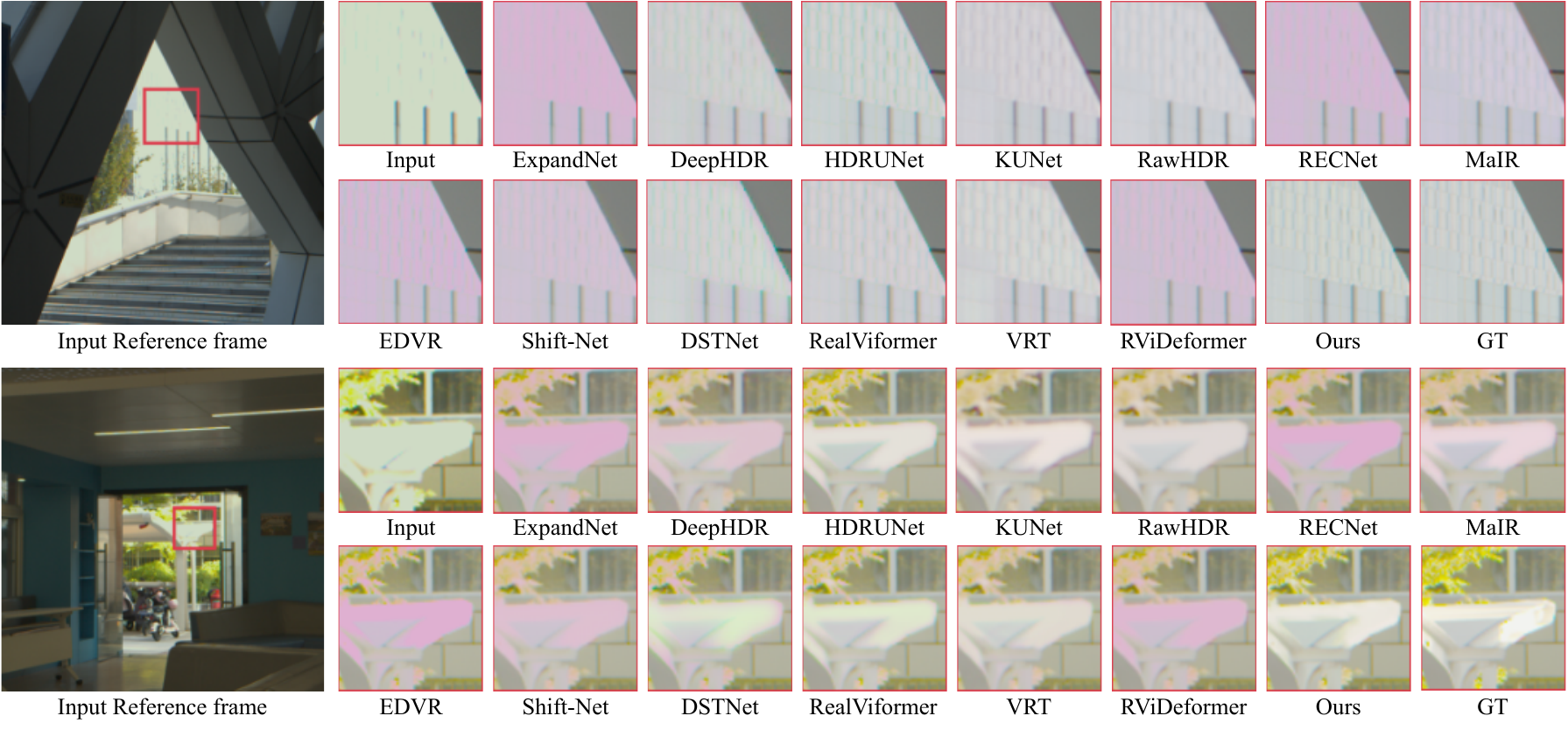}
  \vspace{-1.5em} 
  \caption{Visual comparison of different models in high-light regions. All images are visualized via tone mapping.}
  \label{fig:qualitative_hight} 
\end{figure*}

\begin{figure*}[t]
\centering
  \includegraphics[width=1\textwidth]{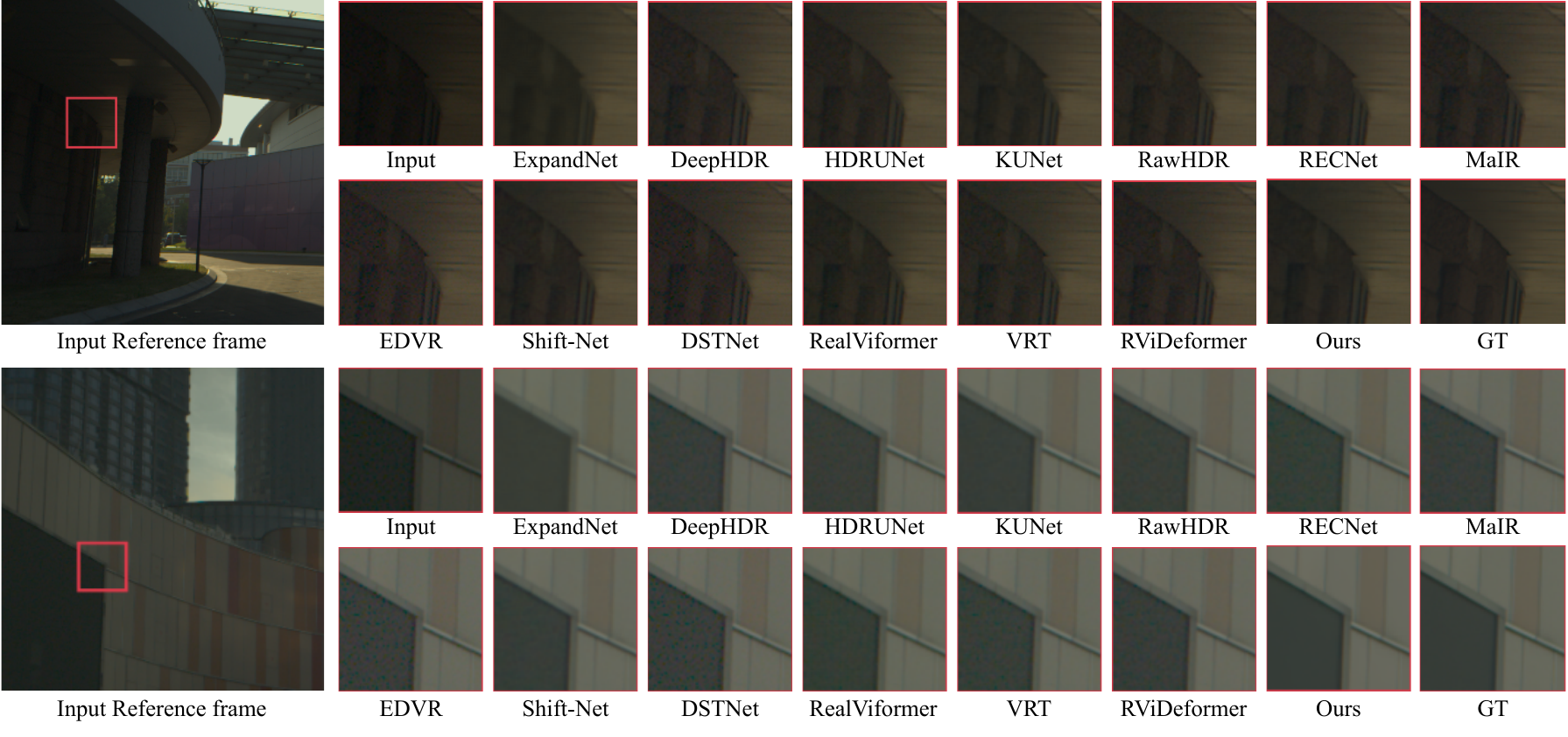}
  \vspace{-1.5em} 
  \caption{Visual comparison of different models in dark regions. All images are visualized via tone mapping.}
  \label{fig:qualitative_dark} 
\end{figure*}

\section{Experiments}
\label{sec:Experiments}

\begin{figure*}[htbp]
  \centering
  \includegraphics[width=1\textwidth, trim=0cm 0.2cm 0cm 0cm, clip]{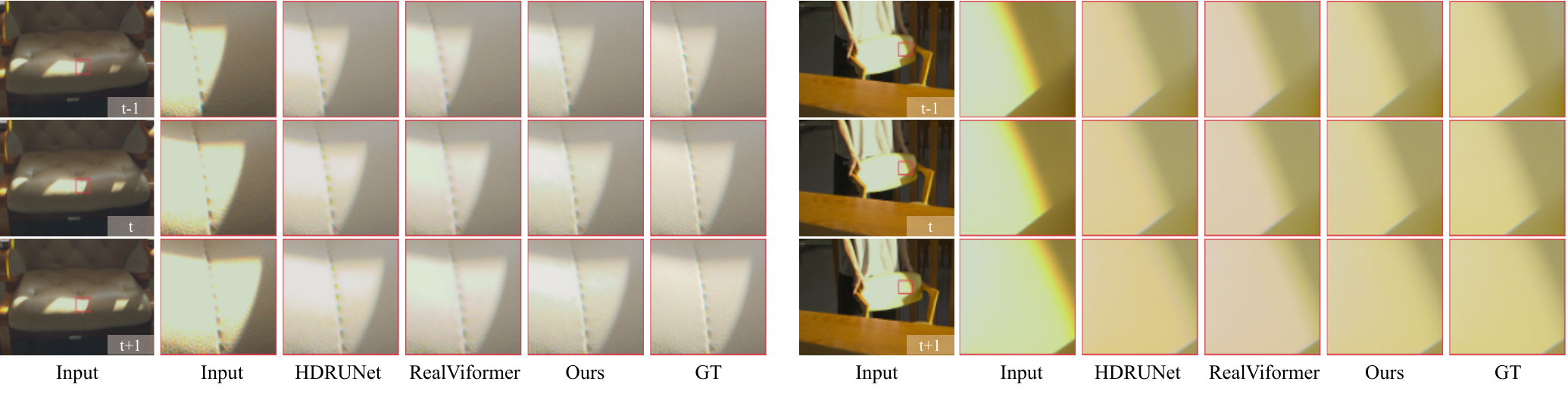}
  \vspace{-1em} 
  \caption{Comparison of highlight recovery on three consecutive frames. Our method leverages temporal exposure complementarity to reconstruct saturated regions, outperforming HDRUNet~\cite{chen2021hdrunet} and RealViformer~\cite{zhang2024realviformer}.}
  \label{fig:Exposure_complementarity}
\end{figure*}

This section presents the experimental setup, including implementation details, comparison methods, and evaluation metrics. We then evaluate our method quantitatively and qualitatively on our RawHDRV dataset, followed by ablation studies to analyze the contribution of each component.

\subsection{Experimental Settings}
\noindent\textbf{Implementation Details.}
Our method is trained on single-exposure (EV0) Raw video sequences, with each sample comprising three consecutive frames and the middle frame as the target. Inputs are normalized in the Raw domain and randomly cropped to \(128 \times 128\) patches. Training batch size is set to $1$, Adam optimizer is used, and initial learning rate is $1\times 10^{-5}$, which is reduced by a factor of $10$ at epoch 50. The entire network is trained for 100 epochs on a single NVIDIA RTX 5070Ti GPU using the PyTorch framework. For fair comparison, all compared methods are retrained on our RawHDRV dataset using the same training settings.

\vspace{0.5em} 
\noindent\textbf{Comparison Methods.}
To comprehensively evaluate our method, we compare against a diverse set of representative models spanning single-image HDR reconstruction, exposure correction, and image and video restoration. Specifically, image restoration methods include ExpandNet~\cite{marnerides2018expandnet}, DeepHDR~\cite{santos2020single}, HDRUNet~\cite{chen2021hdrunet}, KUNet~\cite{wang2022kunet}, RawHDR~\cite{zou2023rawhdr}, RECNet~\cite{liu2024region}, and MaIR~\cite{li2025mair}. Video restoration models include RViDeformer~\cite{yue2025rvideformer}, DSTNet~\cite{pan2023deep}, VRT~\cite{liang2024vrt}, Shift-Net~\cite{li2023simple}, EDVR~\cite{wang2019edvr} and RealViformer~\cite{zhang2024realviformer}. As no public method is specifically designed for single-exposure video HDR reconstruction, this selection covers the most relevant directions and enables multi-faceted validation of our approach in terms of cross-frame consistency, exposure recovery, and detail enhancement.

\noindent\textbf{Evaluation Metrics.}
We use six metrics, including PSNR-L, SSIM-L, PSNR-$\mu$, MS-SSIM, HDR-VDP-2~\cite{mantiuk2011hdr}, and HDR-VQM~\cite{narwaria2015hdr}. PSNR-L and SSIM-L are computed in the logarithmic domain, while PSNR-$\mu$ and MS-SSIM use $\mu$-law tone mapping ($\mu=5000$). HDR-VDP-2 and HDR-VQM are evaluated on linear EXR outputs. Specifically, HDR-VDP-2 is evaluated assuming a 30-inch display at 1.2 m distance, and HDR-VQM measures temporal stability, where lower scores indicate better consistency.

\subsection{Evaluation Results on Our Dataset}
To validate the effectiveness of our method, we conduct comprehensive comparisons on the proposed RawHDRV dataset. The dataset is split into 350 sequences for training and 50 for testing (30 daytime and 20 nighttime scenes). Image-based models~\cite{marnerides2018expandnet,santos2020single,chen2021hdrunet,wang2022kunet,liu2024region,li2025mair,zou2023rawhdr} are applied frame-wise. For those originally designed for 3-channel input~\cite{marnerides2018expandnet,santos2020single,chen2021hdrunet,wang2022kunet,liu2024region,li2025mair,wang2019edvr,pan2023deep,li2023simple,liang2024vrt,zhang2024realviformer}, we adapt the input layer to accept 4-channel Bayer Raw data. All evaluations use identical splits, resolutions, and protocols for fairness.

\vspace{0.5em} 
\noindent\textbf{Quantitative Results.}
Table~\ref{tab:comparison} summarizes quantitative results across six evaluation metrics. Our method consistently achieves the best performance, outperforming the second-best method by 2.54 dB in PSNR-L, 2.39 dB in PSNR-$\mu$, and 1.36 in HDR-VDP-2. This demonstrates superior performance in both pixel fidelity and perceptual HDR realism.

\vspace{0.5em} 
\noindent\textbf{Qualitative Results.}
Fig.~\ref{fig:qualitative_hight} and Fig.~\ref{fig:qualitative_dark} shows qualitative comparisons in typical scenes. Our method accurately recovers details in overexposed and underexposed regions, while others suffer from highlight color shifts and noisy shadows. Fig.~\ref{fig:Exposure_complementarity} validates our exposure complementarity mask-guided restoration: by adaptively fusing reliable information from neighboring frames, our model successfully reconstructs dynamic overexposed regions where competitors fail due to inadequate temporal exploitation.

\begin{figure*}[htbp]
  \centering
  \includegraphics[width=1\textwidth, trim=0cm 0.3cm 0cm 0cm, clip]{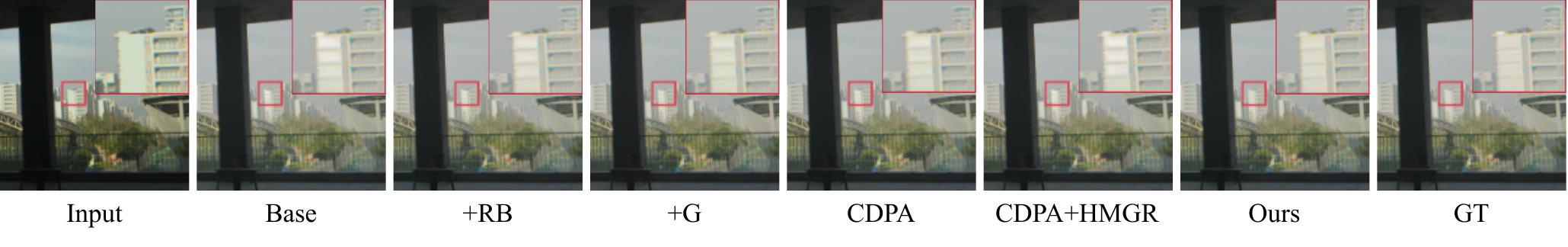}
  \vspace{-1.5em} 
  \caption{Visual ablation study of the proposed components.}
  \label{fig:channel_visual}
\end{figure*}

\begin{figure*}[ht]
  \centering
  \includegraphics[width=1\textwidth]{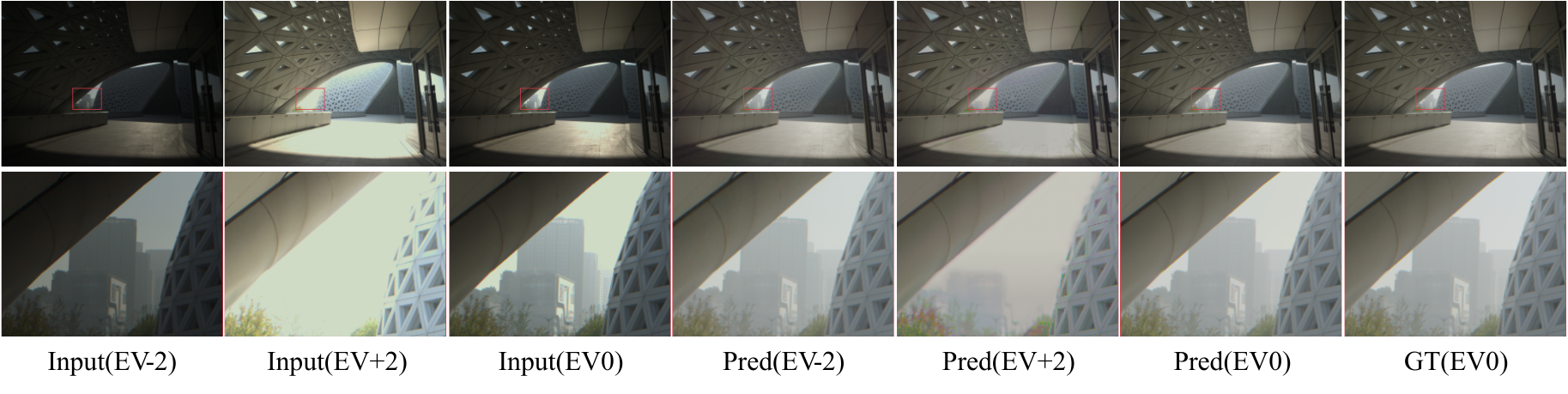}
  \vspace{-1.5em}
  \caption{Visual comparison of results under different exposure values (EV$-$2, EV0, EV$+$2).}
  \label{fig:exposure_visual}
\end{figure*}

\begin{table}[h]
  \centering
  \caption{Ablation study on Channel-Decomposition Parallel Alignment (CDPA) and Mask-Guided Restoration (MGR).}
  \vspace{-0.5em} 
  \label{tab:channel_ablation}
  \small
  \setlength{\tabcolsep}{2.5pt} 
  \resizebox{0.48\textwidth}{!}{%
  \begin{tabular}{@{\hspace{6pt}}l|cccccc@{\hspace{6pt}}}
    \toprule
    Module  & Base & +RB & +G & \makecell{+RB+G\\(CDPA)} & \makecell{CDPA\\+HMGR} & \makecell{CDPA+ECMGR\\(Ours)} \\
    \midrule
    X  & \checkmark & \checkmark & \checkmark & \checkmark & \checkmark & \checkmark \\
    RB &            & \checkmark &            & \checkmark & \checkmark & \checkmark \\
    G  &            &            & \checkmark & \checkmark & \checkmark & \checkmark \\
    HMGR &          &            &            &            & \checkmark &            \\
    ECMGR  &          &            &            &            &            & \checkmark \\
    \midrule
    PSNR-$\mu$$\uparrow$      & 36.49 & 36.72 & 36.78 & 36.84 & 36.48 & \textbf{37.77} \\
    MS-SSIM$\uparrow$      & 0.9902 & 0.9906 & 0.9903 & 0.9903 & 0.9898 & \textbf{0.9916} \\
    HDR-VDP-2$\uparrow$   & 71.87 & 72.10 & 71.89 & 72.03 &  71.80 & \textbf{72.45} \\
    HDR-VQM$\downarrow$  & 0.0080 & 0.0077 & 0.0075 & 0.0076 & 0.0076 & \textbf{0.0066} \\
    \bottomrule
  \end{tabular}%
  }
\end{table}

\subsection{Ablation Studies}
We conduct ablation studies on the RawHDRV dataset with consistent training and evaluation settings, varying only the module or loss to assess each component’s contribution.

\vspace{0.5em} 
\noindent\textbf{Channel-Decomposition Parallel Alignment.}
As shown in Table~\ref{tab:channel_ablation} and Fig.~\ref{fig:channel_visual}, using only the X branch (Base) yields the weakest performance, especially in extreme exposure regions. Adding the RB branch improves highlight recovery, while the G branch enhances shadow detail. The full CDPA configuration (X+RB+G) achieves the best results across all metrics, confirming that channel decomposition aligned with physical characteristics yields complementary gains.

\vspace{0.5em} 
\noindent\textbf{Exposure Complementarity Mask-Guided.}
Building on CDPA, we validate the proposed mask-guided restoration strategy. Replacing standard fusion with a hard binary mask (CDPA+HMGR) introduces color artifacts near saturation boundaries due to abrupt mask transitions. In contrast, our softened mask (CDPA+ECMGR, Ours), obtained by Gaussian blurring the hard mask, produces spatially smooth weighting, effectively suppressing artifacts and achieving the best performance.

\begin{table}[t]
  \centering
  \caption{Quantitative comparison of different loss functions.}
  \vspace{-0.5em} 
  \label{tab:loss_ablation}
  \resizebox{0.48\textwidth}{!}{
  \begin{tabular}{@{\hspace{6pt}}l|ccc@{\hspace{6pt}}}
    \toprule
    Loss configuration & PSNR-$\mu$$\uparrow$ & MS-SSIM$\uparrow$ & HDR-VDP-2$\uparrow$ \\
    \midrule
    $\mathcal{L}_{\text{L1}}$ only & 36.31 & 0.9898 & 71.95 \\
    $\mathcal{L}_{\text{L1}} + \mathcal{L}_{\text{log-L2}}$ & 36.67 & 0.9902 & 71.69 \\
    $\mathcal{L}_{\text{L1}} + \mathcal{L}_{\text{log-L2}} + \mathcal{L}_{\text{m}}$ & \textbf{37.77} & \textbf{0.9916} & \textbf{72.45} \\
    \bottomrule
  \end{tabular}
  }
\end{table}

\begin{figure}[t]
  \centering
  \vspace{-0.5em}
  \includegraphics[width=0.48\textwidth, trim=0cm 0.4cm 0cm 0cm, clip]{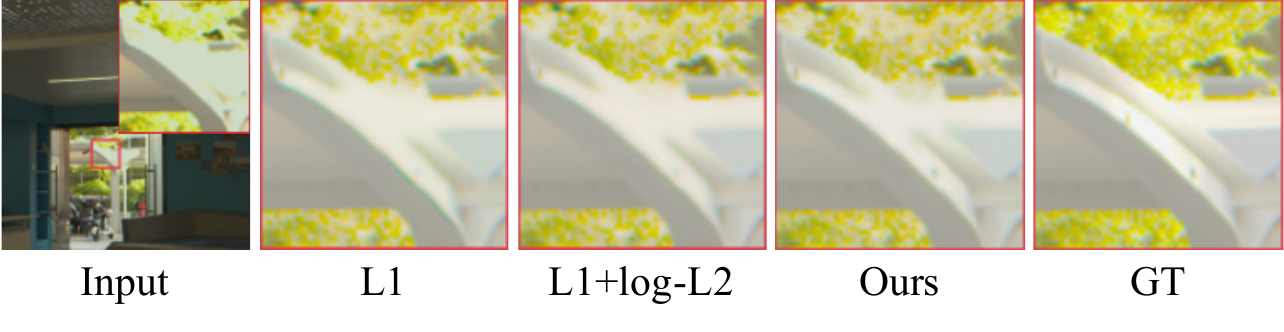}
  \vspace{-0.5em} 
  \caption{Visual comparison of different loss function results.}
  \label{fig:loss_visual}
\end{figure}

\begin{table}[t]
\centering
\vspace{-1em}
\caption{Quantitative results on Sony IMX921 sensor.}
\vspace{-0.5em}
\label{tab:new_data}
\renewcommand{\arraystretch}{0.9}
\resizebox{0.98\columnwidth}{!}{
\begin{tabular}{l|cccc}
\toprule
Method & PSNR-$\mu$$\uparrow$ & MS-SSIM$\uparrow$ & HDR-VDP-2$\uparrow$\\
\midrule
RealViformer~\cite{zhang2024realviformer} & 35.48 & 0.9836 & 70.44 \\
Ours & \textbf{37.42} & \textbf{0.9887} & \textbf{71.25}\\
\bottomrule
\end{tabular}
}
\end{table}

\begin{figure}[t]
  \centering
  \vspace{-0.5em}
  \includegraphics[width=0.48\textwidth]{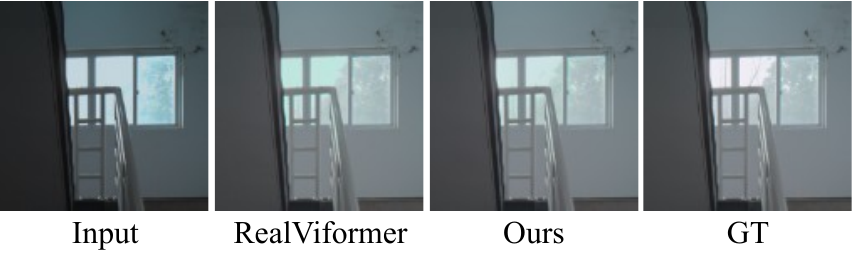}
  \vspace{-0.6em} 
  \caption{Visualization results on the new dataset.}
  \label{fig:new_data} 
\end{figure}

\begin{table}[t]
  \centering
  \caption{Quantitative comparison of reconstruction quality under different input exposure values.}
  \vspace{-0.5em}
  \label{tab:exposure_extended}
  \setlength{\tabcolsep}{1.5pt}
  \resizebox{\linewidth}{!}{
    \begin{tabular}{l|cccc}
      \toprule
      \textbf{Exposure} & \textbf{PSNR-$\mu$}$\uparrow$ & \textbf{MS-SSIM}$\uparrow$ & \textbf{HDR-VDP-2}$\uparrow$ & \textbf{HDR-VQM}$\downarrow$ \\
      \midrule
      $\text{EV}-2$  & 34.79 & 0.9815 & 68.93 & 0.0390 \\
      $\text{EV}+2$  & 24.01 & 0.9628 & 65.34 & 0.1864 \\
      $\text{EV}0$  & \textbf{37.77} & \textbf{0.9916} & \textbf{72.45} & \textbf{0.0066} \\
      \bottomrule
    \end{tabular}
  }
\end{table}

\vspace{0.5em}
\noindent\textbf{Loss Function.}
As shown in Table~\ref{tab:loss_ablation} and Fig.~\ref{fig:loss_visual}, using only $\mathcal{L}_{L1}$ leads to stable convergence but poor highlight recovery. Adding $\mathcal{L}_{\text{log-L2}}$ balances bright and dark regions in the log domain, yielding smoother tone transitions. The full loss, including our Mask-Guided Color Loss $\mathcal{L}_{\text{m}}$, achieves the best metrics, recovering more highlight details and reducing boundary artifacts, confirming the importance of exposure-aware supervision and gradient smoothing.

\subsection{Cross-Sensor Generalization}
To verify generalization capability, we captured 10 additional sequences using the Sony IMX921 sensor for zero-shot testing. Standardized preprocessing (black level subtraction and normalization) minimizes hardware-specific differences, while the shared Bayer structure enables effective transfer of our cross-channel compensation and temporal complementarity strategies. As shown in Table~\ref{tab:new_data} and Fig.~\ref{fig:new_data}, our method achieves 37.42~dB PSNR-$\mu$ without fine-tuning, outperforming RealViformer~\cite{zhang2024realviformer} by 1.94~dB and demonstrating robust generalization across sensors. 

\subsection{Extended Evaluation of RawHDRV}
To verify the architectural versatility of our framework, we conducted extended evaluations on the Raw-HDRV dataset using non-standard exposure inputs. Unlike the main experiments with standard exposure ($\text{EV}0$), we retrained the model to map underexposed ($\text{EV}-2$) and overexposed ($\text{EV}+2$) inputs to the standard HDR ground truth. This extended evaluation addresses real-world scenarios where capture conditions deviate from ideal exposure due to lighting constraints or auto-exposure hysteresis, assessing the framework's potential as a unified solution for exposure correction and dynamic range reconstruction.

As shown in Table~\ref{tab:exposure_extended} and Fig.~\ref{fig:exposure_visual}, the retrained model demonstrates strong adaptability to extreme photometric conditions. For underexposed inputs ($\text{EV}-2$), the model effectively restores standard brightness while suppressing noise. For overexposed inputs ($\text{EV}+2$), despite inevitable information loss in clipped regions, the model leverages exposure-complementary mask-guided spatial priors to infer plausible textures and mitigate color artifacts. These results confirm that RawHDRV can be optimized to handle severe photometric degradations, providing a robust foundation for diverse exposure-correcting HDR applications.
\section{Conclusion}
\label{sec:conclus}

We propose a single-exposure Raw video HDR reconstruction framework. To the best of our knowledge, it is the first to systematically address high dynamic range recovery under uniform exposure conditions. Our method leverages the asymmetric characteristics of Bayer channels through a channel-decomposition parallel alignment mechanism, where Red-Blue and Green streams are processed separately and fused with exposure-aware differential weighting to enhance detail recovery in both highlights and shadows. Additionally, we introduce an exposure complementarity mask-guided restoration module that transfers reliable information from neighboring frames to reconstruct saturated regions, together with a mask-guided color loss that enforces perceptual fidelity and structural smoothness in extreme areas. Evaluated on our newly built RawHDRV dataset, our approach consistently outperforms existing single-image and video restoration methods in highlight and shadow recovery. Future work will focus on improving model efficiency for real-time mobile applications.

\section*{Acknowledgments}
This work was supported by the National Natural Science Foundation of China under Grants No. 62506108 and 62376264, and Brain Science and Brain-like Intelligence Technology---National Science and Technology Major Project under Grant No. 2022ZD0208700.

\bibliographystyle{IEEEtran}
\bibliography{reference}


 




\vfill

\end{document}